\documentclass[letterpaper, 10 pt, conference]{ieeeconf}  

\IEEEoverridecommandlockouts                              

\usepackage{graphics} 
\usepackage{epsfig} 
\usepackage{mathptmx} 
\usepackage{times} 
\usepackage{amsmath} 
\usepackage{amssymb}  
\usepackage{cite}
\usepackage{xcolor}
\usepackage{gensymb}
\usepackage{comment}
\usepackage{multicol,multirow}
\usepackage{mathtools, nccmath}
\usepackage{textcomp, gensymb}
\usepackage{cite}
\usepackage{amsmath,amssymb,amsfonts}
\usepackage{needspace}
\usepackage{microtype}    
\usepackage{bm}           
\usepackage{tabularx}
\usepackage{graphicx}
\usepackage{textcomp}
\usepackage{booktabs}
\usepackage{url}
\usepackage{threeparttable} 
\usepackage{wrapfig}
\usepackage{tabularx}
\usepackage{svg}
\usepackage{makecell}
\usepackage{algorithm}
\usepackage{algpseudocode} 
\usepackage{tabularx}
\usepackage{array}
\newcolumntype{Y}{>{\raggedright\arraybackslash}X}
\title{\LARGE \bf
A Personalized Dynamic Balance Evaluation Paradigm for Hip Exoskeleton-Assisted Walking under Unexpected Ground Perturbations}

\author{Yun Chen, Oluwasegun T. Akinniyi, \IEEEmembership{Student Member,~IEEE}, Qiang Zhang$^{*}$, \IEEEmembership{Member,~IEEE}
\thanks{*This work was supported by the Startup Fund at the University of Alabama. Corresponding author: Qiang Zhang ({\tt\small qiang.zhang@ua.edu}).}
\thanks{Yun Chen, O. T. Akinniyi and Q. Zhang are with the Department of Mechanical Engineering, the University of Alabama, Tuscaloosa, AL 35401, USA.}}

\begin{document}

\maketitle
\thispagestyle{empty}
\pagestyle{empty}

\begin{abstract}
Hip exoskeletons may improve recovery from unexpected gait perturbations, yet personalizing assistance remains difficult because balance is multidimensional and human-in-the-loop experiments are small-sample and noisy. We present a participant-specific composite balance cost that integrates seven biomechanical sub-metrics spanning margin of stability, center-of-mass dynamics, and whole-body angular momentum. The sub-metrics are converted to direction-aligned, dimensionless cost features, and nonnegative fusion weights are learned on the simplex. Coupled with an empirical-Bayes hierarchical model, the learned-composite selector estimates each tested condition's posterior probability of being best, $P(\mathrm{best})$, and a high-probability candidate set with size $K_{0.8}$. The framework was evaluated with three participants walking at 1.1~m/s during unilateral belt-slip perturbations across 46 hip-assistance conditions. In the full-budget analysis ($B=4$ repeats per condition), the selector concentrated 80\% of the posterior probability within 1--5 of 46 conditions, compared with 2--12 for equal-weight fusion and 4--37 for principal component analysis fusion. This smaller candidate set could shorten personalization experiments and limit participants' exposure to repeated perturbations in future studies. Selected-condition trials showed lower observed composite costs than no-torque trials, with nominal $p<0.05$ for P2 and P3. Leave-one-repeat-out refits yielded positive mean held-out rank correlations for all participants and moderate stability of the learned weights and candidate sets. These proof-of-concept results support participant-specific composite balance evaluation for candidate selection in perturbation-based human-in-the-loop experiments.
\end{abstract}

\section{INTRODUCTION}

During everyday walking, individuals frequently encounter sudden disruptions, such as stepping on slippery surfaces, negotiating curbs, or being bumped in crowds. Loss of balance and falls are major drivers of injury risk, reduced mobility, and loss of independence in older adults and individuals with motor impairments \cite{grimmer2019mobility}. The United States (U.S.) Centers for Disease Control and Prevention reports that more than one in four adults aged 65 and older experience falls annually, resulting in millions of emergency department visits and tens of thousands of deaths~\cite{phelan2018fall}. To mitigate fall risks, various mobility aids and rehabilitation programs have been developed. However, conventional tools such as canes and walkers lack the capability to actively respond to sudden balance loss. Wearable lower-limb exoskeletons provide a more proactive solution by monitoring user stability in real time and applying corrective joint torques during stumbles~\cite{monaco2017ecologically}. Nevertheless, most exoskeleton research has focused on reducing the energy cost of steady walking. Few devices are explicitly designed to prevent falls, primarily due to the absence of standardized frameworks for evaluating and optimizing balance recovery~\cite{leestma2024dynamic}.

Reactive balance recovery, referring to rapid, automatic responses to unexpected perturbations, is a key determinant of fall risk and functional mobility \cite{mccrum2017systematic, maki2006control}. Perturbation-based balance training is recognized as an effective paradigm precisely because it targets these rapid reactions under controlled conditions \cite{mccrum2022perturbation}, yet age- and impairment-related deficits in compensatory stepping and limb responses can limit recovery performance. Together, these factors motivate new assistive strategies that improve physical balance recovery while also reducing perceived instability.

Laboratory studies employing treadmills and sudden mechanical perturbations have demonstrated that hip exoskeletons can assist users in recovering balance~\cite{shokouhi2024recovering, monaco2017ecologically}. Existing studies show that hip/pelvis exoskeletons can improve balance recovery after unexpected slip-like disturbances by detecting balance loss and applying counteracting hip torques \cite{monaco2017ecologically}. At the same time, emerging evidence underscores a fundamental control challenge: effective balance assistance requires very fast, well-timed intervention, sometimes faster than physiological responses, and poor timing can negate these benefits \cite{beck2023exoskeletons}. Recent studies also suggest that optimizing exoskeleton assistance for energetics alone may be insufficient to improve reactive stability during gait perturbations, pointing to the need for stability-centric and user-specific personalization strategies \cite{tagliaferri2026systematic}. These findings establish the feasibility of hip-exoskeleton assistance but also expose a fundamental evaluation challenge: no single biomechanical metric comprehensively represents the user’s multidimensional balance-recovery response. This limitation complicates the systematic selection of participant-specific assistance and the future development of adaptive control based on stability-related outcomes.

Typically, determining optimal control strategies remains challenging. Human walking exhibits substantial inter-individual variability, resulting in large differences in the ideal timing and magnitude of exoskeleton assistance~\cite{tagliaferri2026systematic}. Additionally, experimental testing is constrained by human fatigue, limiting the number of torque settings and trials that can be evaluated. This limitation complicates the confident identification of optimal strategies from limited and noisy datasets. Another challenge arises from the fact that ``a good balance objective'' is not a singular, easily quantifiable concept. Researchers employ various metrics to assess balance, such as the center of mass (CoM), margin of stability (MoS), and whole-body angular momentum (WBAM), among others. These metrics capture distinct physical aspects of falls and often yield conflicting information~\cite{shokouhi2024recovering, golyski2025effects}. Consequently, optimizing exoskeleton control based on one control objective may produce strategies that differ substantially from those derived using another objective. The absence of a universal, reliable balance standard impedes the establishment of clear targets for device tuning in human-in-the-loop experiments~\cite{chen2024multi}.

\begin{figure}
    \centering
    \includegraphics[width=1.0\linewidth]{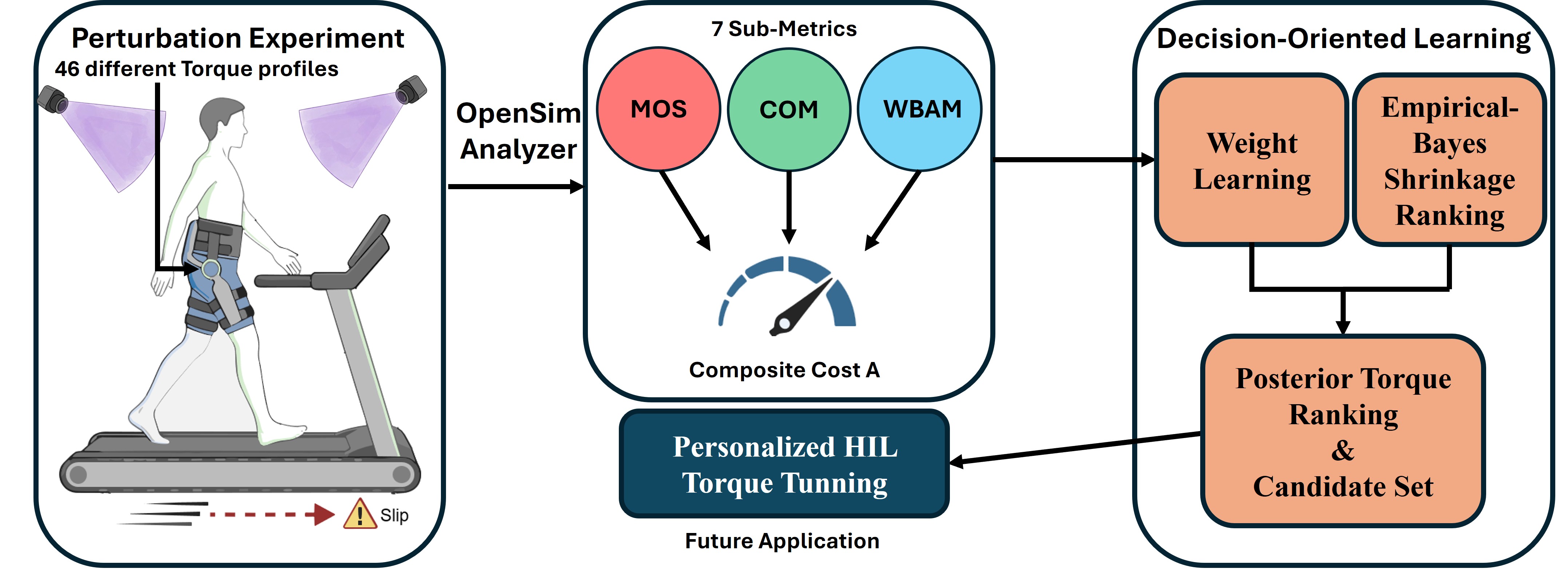}
    \vspace{-5 mm}
    \caption{Schematic overview of the proposed decision-focused composite balance cost and selection framework for small-sample perturbed-walking experiments with hip-exoskeleton assistance.}
    \vspace{-3mm}
    \label{fig:studyoverview}
\end{figure}

To address these challenges, we propose a decision-focused composite balance cost and learned-composite selector for hip-exoskeleton optimization under small-sample conditions (Fig.~\ref{fig:studyoverview}). The composite balance cost combines multiple biomechanical sub-metrics into a unified score. An empirical-Bayes (EB) hierarchical model accommodates the limited and variable data and estimates the posterior probability that each torque condition is best within the tested library. These posterior probabilities provide uncertainty-aware decision summaries for personalizing exoskeleton assistance. In the full-budget analysis ($B=4$ repeats per condition), the learned-composite selector produced more concentrated posterior rankings than the evaluated fusion baselines, supporting future investigation of closed-loop adaptive optimization for assistive robots.
The main contributions of this study include:
1) We formulate a participant-specific composite balance cost from seven direction-aligned MoS-, CoM-, and WBAM-based sub-metrics.

2) We develop a decision-focused selector combining learned fusion weights with EB shrinkage to estimate $P(\mathrm{best})$ and high-probability candidate sets from limited repeats.

3) We evaluate the framework in belt-perturbation experiments with three participants and 46 assistance conditions, comparing equal-weight (EW) and principal component analysis (PCA) fusion and assessing leave-one-repeat-out (LORO) stability.
\vspace{-3mm}

\section{Bilateral Hip Exoskeleton Platform}

\begin{figure}
    \centering
    \includegraphics[width=0.75\linewidth]{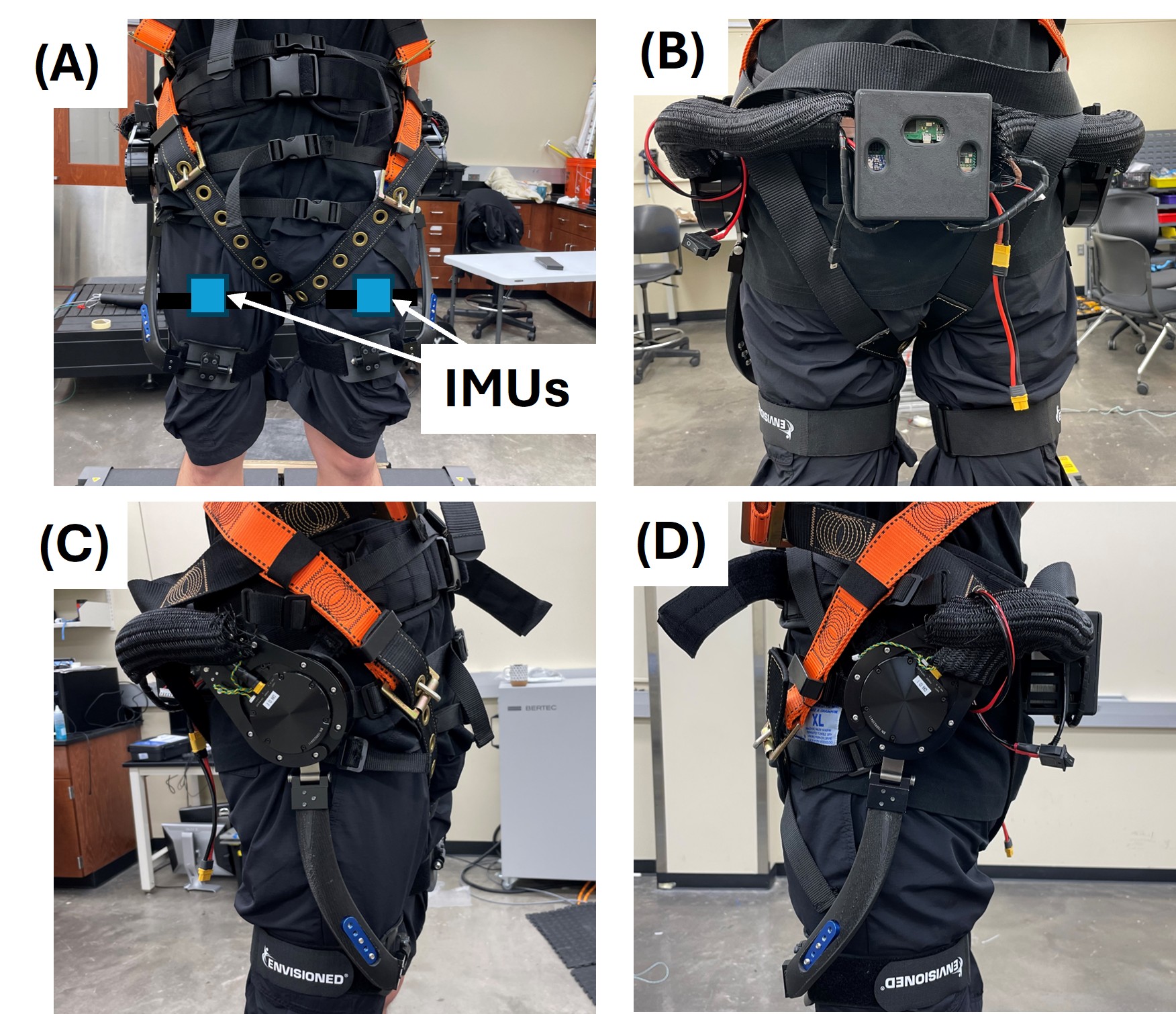}
    \vspace{-2 mm}
    \caption{The lightweight wearable bilateral hip exoskeleton in the current study. (A)-(B) Anterior-posterior views, (C)-(D) Right-left views.}
    \vspace{-3 mm}
    \label{fig:prototype_overview}
\end{figure}

The bilateral hip exoskeleton used in this study is an improved version of our previous design \cite{yu2020quasi}. As shown in Fig.~\ref{fig:prototype_overview}, the platform provides two active hip flexion/extension (HFE) degrees of freedom (DOFs) and two passive hip abduction/adduction (HAA) DOFs. The HFE and HAA ranges of motion are $+135^\circ/-60^\circ$ and $+90^\circ/-60^\circ$, respectively. Each active joint is driven by a CubeMars AK80-9 actuator with a rated continuous torque of $9~\mathrm{N\cdot m}$ and a peak torque of $18~\mathrm{N\cdot m}$. The complete exoskeleton has a mass of 3.6~kg.

Each active joint is instrumented with a flat torque sensor and regulated by closed-loop torque control \cite{nagarajan2016integral}. Two thigh-mounted inertial measurement units (IMUs; LPMS-B2, LP-Research Inc., Japan) measure thigh motion for state-dependent HFE torque commands. The low-level controller communicates with the embedded motor electronics through a controller area network (CAN) bus.

\vspace{-2mm}

\section{Human-in-the-Loop Experiments}
\label{sec:experiments}
\begin{figure*}[htbp]
\centering
  \includegraphics[width=0.99\textwidth]{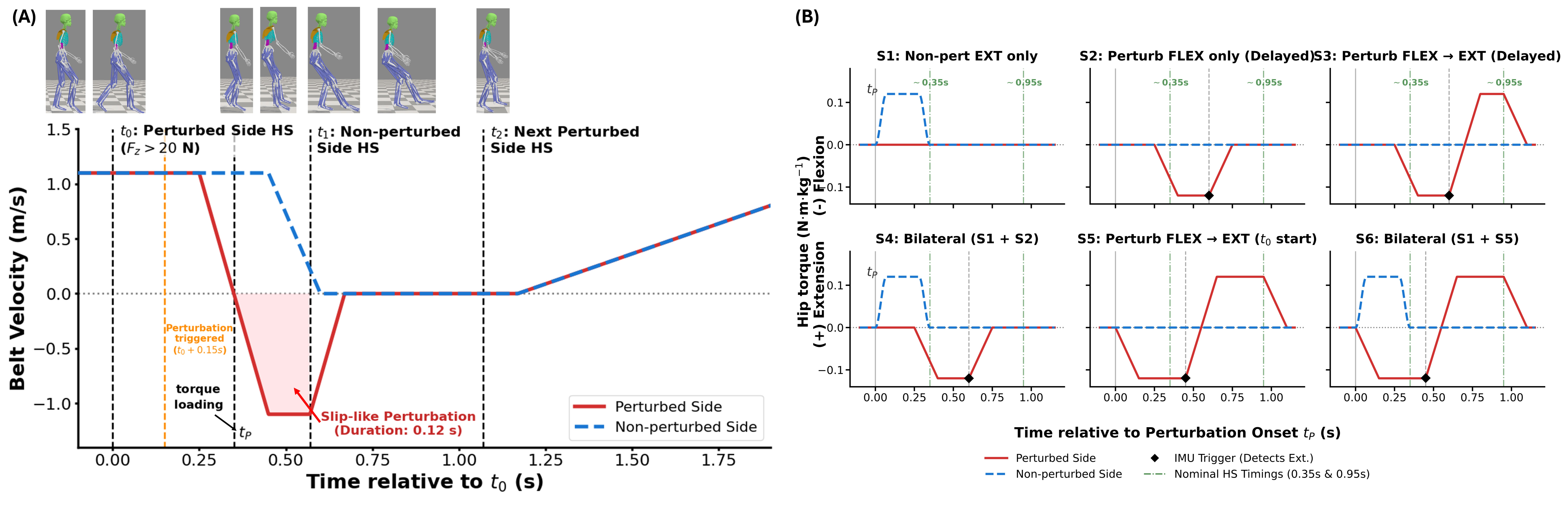}
  \vspace{-4mm}
\caption{Experimental protocol for treadmill-induced slip-like perturbations and hip exoskeleton interventions. (A) Commanded belt velocities for the perturbed (red) and non-perturbed (blue) sides, with gait events and the perturbation sequence marked. (B) Commanded mass-normalized hip torques for scenarios S1--S6. Diamonds mark the IMU-triggered phase switch; positive and negative values denote extension and flexion, respectively.}
\vspace{-5mm}
\label{fig:protocol}
\end{figure*}

Here, non-perturbed (NP) and perturbed (P) denote the two sides; extension (EXT), flexion (FLEX), and flexion-to-extension (F$\rightarrow$E) denote the torque directions.

\begin{table}[t]
  \centering
  \begin{threeparttable}
    \caption{Exoskeleton torque intervention conditions}
    \label{tab:torque_intervention_conditions}

    \scriptsize
    \setlength{\tabcolsep}{3.0pt}
    \renewcommand{\arraystretch}{1.10}

    \begin{tabularx}{\columnwidth}{@{} c l l Y c @{}}
      \toprule
      \textbf{Cond.} & \textbf{NP} & \textbf{P} & \textbf{Swept levels} & \textbf{\#} \\
      \midrule
      S1 & EXT@$t_0$ & --- &
      $A_{np}\in\{0.08,0.12,0.16\}$; $D\in\{250,350\}$ms &
      6 \\

      S2 & --- & FLEX@$t_1$ &
      $A_{p}\in\{0.08,0.12,0.16\}$; $d=350$ms &
      3 \\

      S3 & --- & F$\rightarrow$E@$t_0{+}d$ &
      $A_{p}\in\{0.08,0.12,0.16\}$; $d\in\{250,350\}$ms &
      6 \\

      S4 & EXT@$t_0$ & FLEX@$t_1$ &
      $A_{np},A_{p}\in\{0.08,0.12,0.16\}$; $D\in\{250,350\}$ms &
      18 \\

      S5 & --- & F$\rightarrow$E@$t_0$ &
      $A_{p}\in\{0.08,0.12,0.16\}$ &
      3 \\

      S6$_\mathrm{diag}$ & EXT@$t_0$ & F$\rightarrow$E@$t_0$ &
      $A_p=A_{np}\in\{0.08,0.12,0.16\}$; $D\in\{250,350\}$ms &
      6 \\

      S6$_\mathrm{asym}$ & EXT@$t_0$ & F$\rightarrow$E@$t_0$ &
      $(A_p,A_{np},D)$: four asymmetric settings (Note) &
      4 \\
      \midrule
      \multicolumn{4}{r}{\textbf{Total assisted torque conditions}} & \textbf{46} \\
      \bottomrule
    \end{tabularx}

    \begin{tablenotes}[flushleft]\scriptsize
      \item \textit{Note:}
      $A_p/A_{np}$ are perturbed/non-perturbed-side mass-normalized torque magnitudes (N$\cdot$m$\cdot$kg$^{-1}$);
      $D$ is NP EXT duration;
$d$ is the perturbed-side controller start delay from $t_0$; $t_1\approx t_0{+}350$~ms.
      S6$_\mathrm{asym}$ used $(A_p,A_{np},D\text{ms})=(0.16,0.08,350),(0.16,0.12,350),(0.08,0.16,250),(0.12,0.16,250)$.

    \end{tablenotes}

  \end{threeparttable}
  \vspace{-2mm}
\end{table}

\subsection{Treadmill-triggered slip-like perturbations}
Three young healthy adults (Age: 22$\pm$3~years; height: 175.5$\pm$15~cm; mass: 73.6$\pm$3~kg) without known neurological disorders or mobility deficits participated in the study. The protocol was approved by the Institutional Review Board (IRB; Protocol ID: 23-09-6911), and all participants provided written informed consent.

Participants walked at 1.1~m/s on a force-instrumented split-belt treadmill (Bertec Corp., Columbus, OH, USA). Slip-like anterior--posterior perturbations were applied via a unilateral belt deceleration--reversal profile during stance of the perturbed limb (Fig.~\ref{fig:protocol}A). Heel strike (HS) and toe-off (TO) were detected online from the vertical ground reaction force ($F_z$).

To target triggering perturbation at mid single-stance, the command was issued 0.15~s after perturbed-side HS ($t_0$), whereas perturbation onset ($t_P$) was defined as the belt-velocity zero-crossing and subsequent reverse acceleration (Fig.~\ref{fig:protocol}A); separating command issuance from physical onset accommodates treadmill control and communication latency. For offline alignment, $t_1$ denotes the first contralateral HS after $t_P$ and $t_2$ denotes the next perturbed-side HS; outcome measures were computed over the two-step response window $[t_0, t_2]$ (Fig.~\ref{fig:protocol}A). Following the perturbation, both belts were braked to 0 to provide an approximately stationary support surface, and walking resumed by re-accelerating to 1.1~m/s 0.10~s after $t_2$. To reduce anticipation, the perturbation side was pseudo-randomized, and a random waiting period of 5--8 gait cycles was enforced before enabling the next trigger, while keeping perturbations balanced between sides.
\vspace{-1mm}
\subsection{Hip exoskeleton torque interventions}
We tested six exoskeleton torque-intervention scenarios (S1--S6) with separate no-torque reference trials. Fig.~\ref{fig:protocol}B shows the commanded hip-torque profiles and timing conventions, including the flexion/extension sign and IMU-triggered phase switch. Table~\ref{tab:torque_intervention_conditions} summarizes the scenario-specific parameter sweeps and 46 resulting unique torque conditions.

Each participant completed six sessions corresponding to S1--S6, comprising 184 assisted perturbation trials and 40 no-torque reference trials. Total recorded durations were 28.7, 30.7, and 29.7~min for P1, P2, and P3, respectively. Session and torque-condition orders were randomized. Each assisted condition was applied in four consecutive trials, with each trial constituting one repeat. One no-torque reference trial was inserted between condition blocks.
\vspace{-1mm}
\subsection{Data collection and preprocessing}

During the experiments, the exoskeleton system was supervised through a graphical user interface (GUI) running on the host personal computer (PC). The GUI communicated with the remote computer controlling the exoskeleton and simultaneously maintained a real-time User Datagram Protocol (UDP) connection with the force-instrumented split-belt treadmill (Bertec, USA) to stream bilateral belt velocities as well as the three-dimensional ground reaction forces and moments from the embedded force plates. These signals were used for online gait event detection and, according to the predefined protocol, to trigger exoskeleton torque assistance and send treadmill commands for slip-like perturbations.

Kinematic data were collected using an optical motion capture system (Vicon, Vicon Motion Systems, UK) with a full-body marker set placed on major bony landmarks. Three-dimensional marker trajectories were recorded, zero-phase low-pass filtered at 6~Hz, and imported into OpenSim. A full-body musculoskeletal model previously described in the literature~\cite{rajagopal2016full} was scaled to each participant, and inverse kinematics was performed to obtain joint angle trajectories. Based on these results, OpenSim analysis tools were used to compute WBAM during the perturbation interval and the three-dimensional kinematics of both heel joint centers.
\section{Composite Balance Cost Formulation and Weight Optimization}
\vspace{-2mm}
\label{sec:methods}

\subsection{Biomechanical sub-metrics and composite balance cost construction}

Using these processed kinematic outputs, we compute seven biomechanical sub-metrics over the two-step recovery window $[t_0,t_2]$, where $t_0$ and $t_2$
denote the analysis-window onset and end events on the experimental timeline defined in Section~\ref{sec:experiments}.
These sub-metrics capture (i) extrapolated center of mass (XCoM)/base of support (BoS) stability margin, (ii) linear CoM responses, and (iii) sagittal-plane
whole-body angular momentum (WBAM) regulation. We denote the anteroposterior (AP) and vertical CoM components by
$x_{\mathrm{AP}}$ and $y_{\mathrm{CoM}}$, respectively. In implementation, $x_{\mathrm{AP}}$ is taken from the
treadmill progression axis in the OpenSim ground frame, with negative values indicating posterior motion.
The margin of stability (MoS) characterizes endpoint stability at the second recovery heel strike. The CoM
displacement and acceleration sub-metrics characterize transient translational responses, whereas the WBAM sub-metrics
characterize accumulated and range-based rotational responses. We selected these three functional categories to
represent complementary single-trial outcomes aligned with the AP perturbation and sagittal-plane assistance in
this study. Frontal-plane balance remains relevant but was outside this task-specific sub-metric set; foot placement is
represented indirectly through the base-of-support (BoS) boundary in MoS, whereas step-timing, step-width
variability, and foot-placement predictability were not included because the analysis targeted individual recovery
trials with four repeats per condition \cite{wu2025detecting}.
Table~\ref{tab:balance_submetrics} summarizes the sub-metric names, evaluation windows, physical directions, and
reference definitions.

\begin{table}[t]
\centering
\caption{Sub-metrics used to construct the composite balance cost}
\label{tab:balance_submetrics}
\scriptsize
\setlength{\tabcolsep}{2.5pt}
\renewcommand{\arraystretch}{1.05}
\begin{tabular}{@{}clccc@{}}
\hline
$k$ & Sub-metric $m_k$ & Window & Better & Ref.\\
\hline
$1$ & $\mathrm{MoS}_{\mathrm{post}}(t_2)$ & $t_2$ & $\uparrow$ & \cite{hof2005condition,young2012dynamic}\\
$2$ & peak~$|a_{y,\mathrm{CoM}}|$ & $[t_0,t_2]$ & $\downarrow$ & \cite{van2020effect}\\
$3$ & $\Delta x^{\max}_{\mathrm{CoM,post}}$ & $[t_0,t_2]$ & $\downarrow$ & \cite{matjavcic2019influence}\\
$4$ & abs.~sagittal iWBAM dev. (SD) & $[t_0,t_2]$ & $\downarrow$ & \cite{leestma2023linking,liu2020asymmetric,acasio2025osseointegrated}\\
$5$ & peak~$|a_{\mathrm{AP},\mathrm{CoM}}|$ & $[t_0,t_2]$ & $\downarrow$ & \cite{van2020effect}\\
$6$ & sag.~WBAM range change (\%) & $[t_0,t_2]$ & $\downarrow$ & \cite{leestma2023linking,liu2020asymmetric}\\
$7$ & peak~$|\Delta y_{\mathrm{CoM}}|$ & $[t_0,t_2]$ & $\downarrow$ & \cite{matjavcic2019influence}\\
\hline
\end{tabular}
\vspace{-5mm}
\end{table}

Briefly, $\mathrm{MoS}_{\mathrm{post}}(t_2)$ is evaluated at the recovery-step heel strike $t_2$ using the XCoM/BoS
framework \cite{hof2005condition,young2012dynamic}, where the posterior BoS boundary is defined by the minimum
AP coordinate of the bilateral heel joint centers at $t_2$. The CoM-kinematics sub-metrics summarize transient linear
responses over $[t_0,t_2]$ via the peak absolute CoM accelerations in the vertical and AP directions, the maximum
posterior CoM excursion $\max\!\left(0,\max_{t\in[t_0,t_2]}\!\big(x_{\mathrm{ref}}-x_{\mathrm{AP}}(t)\big)\right)$, and the peak
absolute vertical CoM displacement $\max_{t\in[t_0,t_2]}\!|y_{\mathrm{CoM}}(t)-y_{\mathrm{ref}}|$, where
$x_{\mathrm{ref}}$ and $y_{\mathrm{ref}}$ are the mean CoM coordinates over the 100 ms interval preceding $t_0$
\cite{van2020effect,matjavcic2019influence}. For WBAM regulation, we compute WBAM about the whole-body CoM
\cite{leestma2023linking,liu2020asymmetric} and retain only the sagittal-plane component $H_{\mathrm{sag}}$, i.e.,
the component about the mediolateral axis. The signal is normalized by $mvh$, where $m$ is body mass, $v$ is
walking speed, and $h$ is body height. Specifically, we use (i) the absolute integrated sagittal WBAM,
$\mathrm{iWBAM}_{\mathrm{sag}}=\int_{t_0}^{t_2}|H_{\mathrm{sag}}(t)|\,dt$, expressed as its deviation in standard
deviations from the mean over four pre-perturbation gait cycles, and
(ii) the percentage change in sagittal WBAM range,
$100\,(R_{\mathrm{post}}-\bar{R}_{\mathrm{pre}})/\bar{R}_{\mathrm{pre}}$,
where $R_{\mathrm{post}}=\max_{t\in[t_0,t_2]}H_{\mathrm{sag}}(t)-\min_{t\in[t_0,t_2]}H_{\mathrm{sag}}(t)$ and
$\bar{R}_{\mathrm{pre}}$ is the mean sagittal-WBAM range over the same four pre-perturbation gait cycles.

To build the composite balance cost, we use a participant-specific pooled reference dataset, $\mathcal{D}_{\mathrm{NT}}$, containing all no-torque
reference trials. Let the seven raw biomechanical sub-metrics $m_k$, $k=1,\dots,7$, follow the ordering in Table~\ref{tab:balance_submetrics}.
For each $m_k$, we compute a robust location $\mu_k$ and scale $s_k$ over $\mathcal{D}_{\mathrm{NT}}$ using the median and interquartile
range (IQR). If the IQR is nonfinite or no greater than $10^{-9}$, $s_k$ is replaced by $1.4826$ times the median
absolute deviation (MAD); any remaining nonfinite or no greater than $10^{-9}$ value is set to $10^{-9}$
\cite{rousseeuw1993alternatives}. The dimensionless cost features are then defined as
\begin{equation}
z_k=d_k\frac{m_k-\mu_k}{s_k},\qquad
d_1=-1,\quad d_{2\ldots7}=+1,
\end{equation}
which enforces a ``larger-is-worse'' convention. The resulting standardized feature vector is
$\mathbf{z}=[z_1,z_2,z_3,z_4,z_5,z_6,z_7]^{\top}$.
Although some sub-metrics are already normalized, all seven cost features are standardized using the same no-torque reference trials so that no single sub-metric dominates the fusion purely by scale.

We define a nonnegative weight vector $\mathbf{w}=[w_1,\dots,w_7]^{\top}$ with $w_k\ge 0$ and
$\sum_{k=1}^{7}w_k=1$. The composite balance cost is
\begin{equation}
A(\mathbf{z};\mathbf{w})=\sum_{k=1}^{7}w_kz_k,
\end{equation}
where smaller $A$ indicates better overall balance in the composite sense.

\subsection{EB inference and decision-focused learning}

Because each torque condition has only a few repeats, ranking conditions by the sample mean of $A$ can be
unstable. We therefore combine (i) EB hierarchical shrinkage for condition-level ranking and
(ii) decision-focused learning of the fusion weights.

\subsubsection{EB hierarchical shrinkage for torque-condition ranking}

Let $u\in\mathcal{U}=\{1,\dots,U\}$ index torque conditions and $r\in\{1,\dots,R_u\}$ index repeats. For a given
weight vector $\mathbf{w}$, the composite balance cost for repeat $r$ is $A_{u,r}=A(\mathbf{z}_{u,r};\mathbf{w})$. We use the
normal--normal hierarchical model
\begin{equation}
A_{u,r}\mid \theta_u,\sigma^2 \sim \mathcal{N}(\theta_u,\sigma^2), \qquad
\theta_u\mid\mu,\tau^2 \sim \mathcal{N}(\mu,\tau^2),
\label{eq:eb_model}
\end{equation}
where $\theta_u$ is the latent mean cost for condition $u$. The hyperparameters $(\mu,\tau^2,\sigma^2)$ are
estimated by a plug-in EB moment procedure: $\mu$ is the mean of the condition sample means, $\sigma^2$ is the
pooled within-condition variance, and $\tau^2$ is the between-condition variance of the sample means after
subtracting the mean sampling variance $\sigma^2/R_u$. Both variance estimates use a positive floor of $10^{-9}$.
This yields a Gaussian posterior
$\theta_u\mid\mathcal{D}$ with closed-form mean $m_u$ and variance $v_u$, where $m_u$ shrinks noisy condition means
toward the global mean when $R_u$ is small \cite{laird1989ebranking}.

\subsubsection{Decision summaries: $P(\mathrm{best})$ and a candidate set}

From $\{\theta_u\mid\mathcal{D}\}_{u\in\mathcal{U}}$, we compute two decision summaries. First, the posterior
probability that condition $u$ is best among the tested conditions is
\begin{equation}
P_u=\Pr\!\left(u=\operatorname*{arg\,min}_{v\in\mathcal{U}}\theta_v \,\middle|\, \mathcal{D}\right).
\label{eq:pbest_def}
\end{equation}
We define the selected (top-ranked) condition as $\hat u=\arg\max_{u\in\mathcal{U}}P_u$. Its posterior probability of being best is
\begin{equation}
P(\mathrm{best})=P_{\hat u}=\max_{u\in\mathcal{U}}P_u.
\label{eq:pbest_summary}
\end{equation}
In practical terms, a larger $P(\mathrm{best})$ indicates greater confidence that the selected condition is best among the tested conditions.
We estimate $\{P_u\}$ by joint Monte Carlo sampling from the condition posteriors and counting the minimizing
index across samples \cite{russo2020bestarm}.

Second, we form a high-probability candidate set $\mathcal{C}_{0.8}$ by sorting $\{P_u\}$ in descending order and
taking the smallest set whose cumulative probability mass reaches 0.8
\cite{gupta1985bayespstar}. Its size is denoted by $K_{0.8}=|\mathcal{C}_{0.8}|$, where smaller $K_{0.8}$
means that fewer plausible conditions are needed to capture 80\% of the posterior probability, indicating less selection ambiguity.

\subsubsection{Decision-focused learning of fusion weights}

We learn $\mathbf{w}$ on the simplex by optimizing the posterior decision summaries. For each $\mathbf{w}$,
we compute $\{A_{u,r}\}$, fit the EB model in \eqref{eq:eb_model}, estimate
$\mathbf{P}=[P_1,\dots,P_U]^{\mathsf T}$ and $(\mathcal{C}_{0.8},K_{0.8})$, and evaluate
\begin{equation}
\begin{aligned}
J(\mathbf{w})=\;&
\alpha_K K_{0.8}
+\alpha_P\!\left(1-P(\mathrm{best})\right)
+\alpha_H H(\mathbf{P})
+\lambda\,\Phi_{\mathrm{guard}}(\mathbf{w}),
\end{aligned}
\label{eq:J_def}
\end{equation}
where $H(\mathbf{P})=-\sum_u P_u\log P_u$ is the posterior entropy. To define the guard term, let
$\bar z_{u,k}$ be the mean of cost feature $k$ for condition $u$, let $b_k$ be the mean of the three smallest values
among $\{\bar z_{u,k}\}_{u=1}^{U}$, and let $\hat u(\mathbf{w})$ denote the selected condition under $\mathbf{w}$. With
$[x]_+=\max(x,0)$, define $r_k(\mathbf{w})=[\bar z_{\hat u(\mathbf{w}),k}-b_k]_+$,
$\bar r(\mathbf{w})=\frac{1}{7}\sum_{k=1}^{7}r_k(\mathbf{w})$, and
$r_{\max}(\mathbf{w})=\max_k r_k(\mathbf{w})$. The guard term is
\begin{equation}
\begin{aligned}
\Phi_{\mathrm{guard}}(\mathbf{w})={}&
[\bar r(\mathbf{w})-\bar r(\mathbf{w}_{\mathrm{EW}})-\delta]_+\\
&+[r_{\max}(\mathbf{w})-r_{\max}(\mathbf{w}_{\mathrm{EW}})-\delta]_+,
\end{aligned}
\label{eq:guard_def}
\end{equation}
where \(\mathbf{w}_{\mathrm{EW}}=[1/7,\ldots,1/7]^{\top}\). We used
$\alpha_K=0.25$, $\alpha_P=1$, $\alpha_H=0.05$, $\lambda=10$, and $\delta=0$. We do not impose any additional
explicit regularization on $\mathbf{w}$, allowing uninformative or redundant sub-metrics to receive near-zero weights.

The learned fusion weights are defined by the constrained optimization
\begin{equation}
\begin{aligned}
\mathbf{w}^{*}\in{}&\operatorname*{arg\,min}_{\mathbf{w}\in\mathbb{R}^{7}}J(\mathbf{w})\\
\mathrm{subject\ to}\quad
&w_k\geq 0,\quad k=1,\ldots,7,\qquad
\sum_{k=1}^{7}w_k=1.
\end{aligned}
\label{eq:wstar_def}
\end{equation}

Because $J(\mathbf{w})$ is non-smooth and Monte Carlo-estimated, we optimize it over the simplex using the derivative-free cross-entropy method (CEM) \cite{deboer2005cem}, summarized in Algorithm~\ref{alg:cem_weights}. CEM used 10 starts, population size $N=200$, elite size $N_e=20$, $T=18$ iterations, $\sigma_0=1$, smoothing factor $\gamma=0.7$, $\sigma_{\min}=0.2$, and covariance regularizer $\eta=10^{-6}$. Final summaries used 100,000 posterior draws.

\begin{algorithm}[t]
\caption{CEM optimization of the learned fusion weights}
\label{alg:cem_weights}
\scriptsize
\begin{algorithmic}[1]
\Require Trial-level sub-metrics, $N,N_e,T,\sigma_0,\gamma,\sigma_{\min},\eta$
\State Initialize the best solution across starts
\For{each independent start}
  \State $\boldsymbol{\mu}\gets\mathbf{0}$, $\boldsymbol{\Sigma}\gets\sigma_0^2\mathbf{I}$
  \For{$t=1,\ldots,T$}
    \State Draw $N$ logits; map each to the simplex by softmax
    \State Evaluate $J(\mathbf{w})$ and retain the $N_e$ lowest-cost samples
    \State Smooth the elite mean and covariance by $\gamma$
    \State Regularize by $\eta$; floor covariance eigenvalues at $\sigma_{\min}^2$
  \EndFor
  \State Update the best solution across starts
\EndFor
\State \Return the weights $\mathbf{w}^{*}$ with the lowest $J$ across starts

\end{algorithmic}
\end{algorithm}

The fusion baselines are EW fusion and PCA fusion. EW uses $\mathbf{w}_{\mathrm{EW}}=[1/7,\ldots,1/7]^{\top}$. PCA fusion retains the minimum number of components explaining at least 85\% of the variance, aligns each component's sign with EW fusion, and combines the sign-aligned scores using normalized explained-variance weights. As a post-hoc diagnostic, we also report a single-metric comparator. The seven single-metric EB selectors are evaluated separately, and the sub-metric with the largest $P(\mathrm{best})$ is retained.

\subsection{Evaluation protocol and statistical analysis}

Analyses were conducted separately for each participant. The full-budget analysis ($B=4$ repeats per condition)
evaluated the learned-composite selector and both fusion baselines over the 46-condition library. We reported
the selected condition, $P(\mathrm{best})$, and $K_{0.8}$. No-torque reference trials provided participant-specific
standardization statistics and the reference for biomechanical comparisons.

Repeat-level stability was assessed with four LORO refits for each participant. In fold $r$,
the $r$th repeat was withheld from all 46 torque conditions, yielding 138 training repeats and 46 held-out
repeats. The CEM weights, PCA fusion model, EB condition summaries, and candidate sets were recomputed from
the three training repeats. EW fusion retained its predefined weights, and its EB summaries were recomputed on the
same training fold. Normalization parameters were estimated from the no-torque reference trials and held fixed across
the four folds.

For each method, held-out ranking consistency was quantified by the Spearman correlation between the 46 training-fold
posterior condition means and their corresponding composite balance costs in the held-out repeat. For the learned-composite selector,
weight stability in fold $r$ was quantified by
\begin{equation}
S_w^{(r)}=
\frac{{\mathbf{w}^{(-r)}}^{\mathsf T}\mathbf{w}^{\mathrm{full}}}
{\|\mathbf{w}^{(-r)}\|_2\,\|\mathbf{w}^{\mathrm{full}}\|_2},
\label{eq:loro_weight_cosine}
\end{equation}
where $\mathbf{w}^{(-r)}$ is the refitted weight vector and $\mathbf{w}^{\mathrm{full}}$ is the learned fusion-weight vector from the full-budget analysis ($B=4$ repeats per condition).
Selection agreement indicated whether the refitted selected condition matched the selected condition from the full-budget analysis. Selection retention
indicated whether the condition selected in the full-budget analysis remained in the refitted $\mathcal{C}_{0.8}$ candidate set. Spearman
correlations and $S_w^{(r)}$ were averaged across the four folds; agreement and retention were reported as counts out of four.

Separately, we compared composite balance costs between selected-condition trials ($n=4$) and no-torque reference trials ($n=40$)
for each participant using a two-sided Welch $t$-test. We report both means, their difference (selected minus reference),
a Welch 95\% confidence interval, Hedges' $g$, and the nominal $p$ value. Negative differences favor the selected
condition. Because the same full-budget trials were used to learn the weights, select the condition, and
estimate its effect, these participant-specific comparisons were treated as exploratory post-selection analyses;
no multiplicity adjustment was applied.
\begin{figure*}[!t]
\centering
  \includegraphics[width=0.99\textwidth]{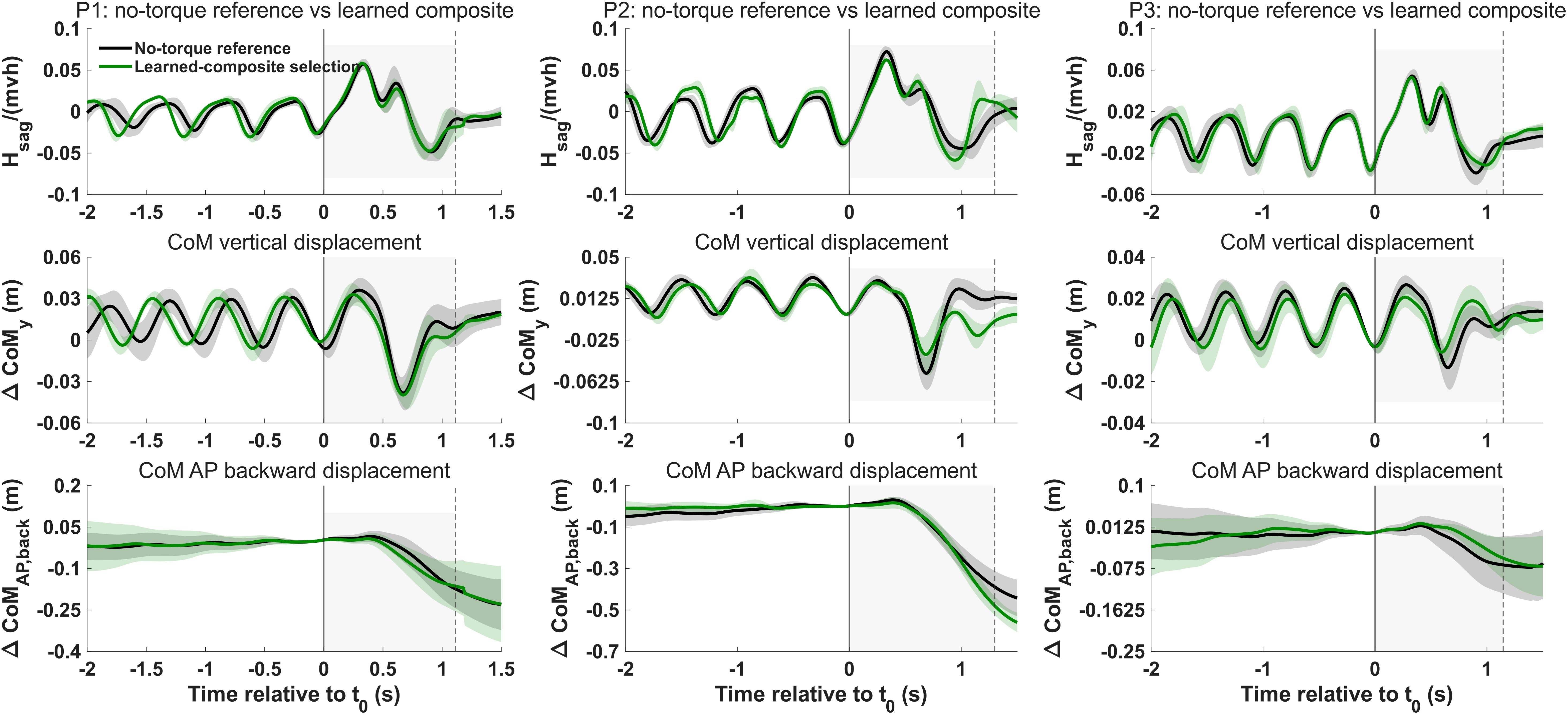}
  \vspace{-3mm}
\caption{Event-aligned recovery time sequences for no-torque reference trials (black) and selected-condition trials (green). Shaded bands denote mean $\pm$ SD across trials.}
\vspace{-2mm}
\label{bio1}
\end{figure*}
\section{RESULTS AND DISCUSSION}
\label{sec:results}

\begin{figure*}[!t]
\centering
  \includegraphics[width=0.99\textwidth]{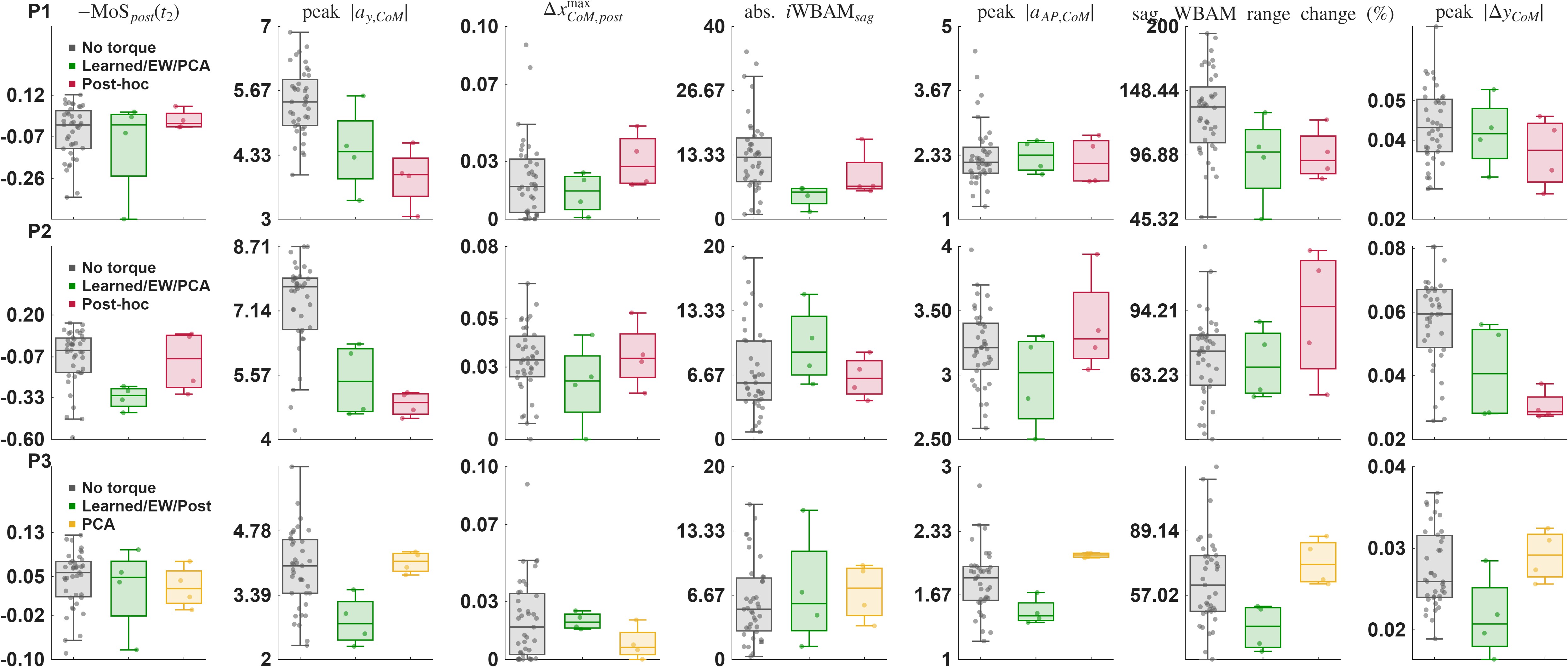}
  \vspace{-3mm}
\caption{Distributions of the seven biomechanical sub-metrics for no-torque reference trials and trials from conditions selected under the full budget by the learned-composite selector, EW and PCA fusion baselines, and single-metric comparator (P1--P3). Methods selecting the same condition are merged in the legend. For display, the first column is $-\mathrm{MoS}_{\mathrm{post}}(t_2)$ so all columns follow a lower-is-better convention. Box plots show the median and interquartile range (whiskers: 1.5$\times$IQR), with individual trials overlaid. The single-metric comparator uses peak $|a_{y,\mathrm{CoM}}|$ for P1 and peak $|\Delta y_{\mathrm{CoM}}|$ for P2 and P3.}
\vspace{-5mm}
\label{bio2}
\end{figure*}

The results are organized around the selected torque conditions and their biomechanical outcomes, the concentration of the posterior rankings in the full-budget analysis, and the repeat-level stability of the learned weights and selected conditions.

\subsection{Selected torque conditions and biomechanical outcomes}
\label{subsec:results_biomech}

In the full-budget analysis ($B=4$ repeats per condition), the selected condition for P1 was condition 30 in S4: non-perturbed-side extension at $A_{np}=0.16~\mathrm{N\,m\,kg^{-1}}$, perturbed-side flexion at $A_p=0.12~\mathrm{N\,m\,kg^{-1}}$, and $D=350$~ms. For P2, condition 5 in S1 used non-perturbed-side extension at $A_{np}=0.12~\mathrm{N\,m\,kg^{-1}}$ and $D=350$~ms. For P3, condition 8 in S2 used delayed perturbed-side flexion at $A_p=0.12~\mathrm{N\,m\,kg^{-1}}$ and $d=350$~ms. EW fusion matched the learned-composite selection for all participants. PCA fusion matched it for P1 and P2 but selected condition 36 for P3. The single-metric comparator differed for P1 and P2 and matched for P3.

Fig.~\ref{bio1} shows event-aligned time sequences of sagittal WBAM, vertical CoM displacement, and AP backward CoM displacement during the perturbation-recovery window. The responses were participant- and signal-specific, with CoM and WBAM differences emerging at different stages of recovery. The selected conditions were associated with distinct recovery signatures across participants. The composite balance cost combines these signals at the trial level and assigns their contributions through participant-specific weights.

In Fig.~\ref{bio2}, the single-metric comparator selected different conditions from the learned-composite selector for P1 and P2. The P3 selections coincided, so their distributions overlapped. The P1 and P2 differences show that one sub-metric can prioritize another region of the torque library.

Across all three participants, the selected-condition trials had a lower observed mean composite balance cost than the no-torque reference trials. For P1, the mean was $-0.755$ for selected-condition trials and $-0.030$ for no-torque reference trials, with a difference of $-0.726$ (Welch 95\% confidence interval (CI) $[-1.548,\,0.096]$, Hedges' $g=-1.805$, nominal $p=0.068$). For P2, the corresponding means were $-1.092$ and $-0.141$, with a difference of $-0.951$ (95\% CI $[-1.654,\,-0.248]$, $g=-1.698$, nominal $p=0.020$). For P3, the means were $-0.657$ and $0.109$, with a difference of $-0.765$ (95\% CI $[-1.324,\,-0.206]$, $g=-1.391$, nominal $p=0.018$). Nominal $p<0.05$ for P2 and P3; P1's confidence interval crossed zero. The common direction of the mean differences indicates that the selected conditions occupied a lower-cost region of the tested library, with different levels of precision across participants. These findings provide preliminary evidence that the participant-specific torque conditions selected by the learned-composite selector may improve the overall stability-related response to slip-like perturbations relative to no-torque reference trials.

\subsection{Posterior decision concentration relative to fusion baselines}
\label{subsec:results_decision}

\begin{figure}[!t]
\centering
  \includegraphics[width=0.49\textwidth]{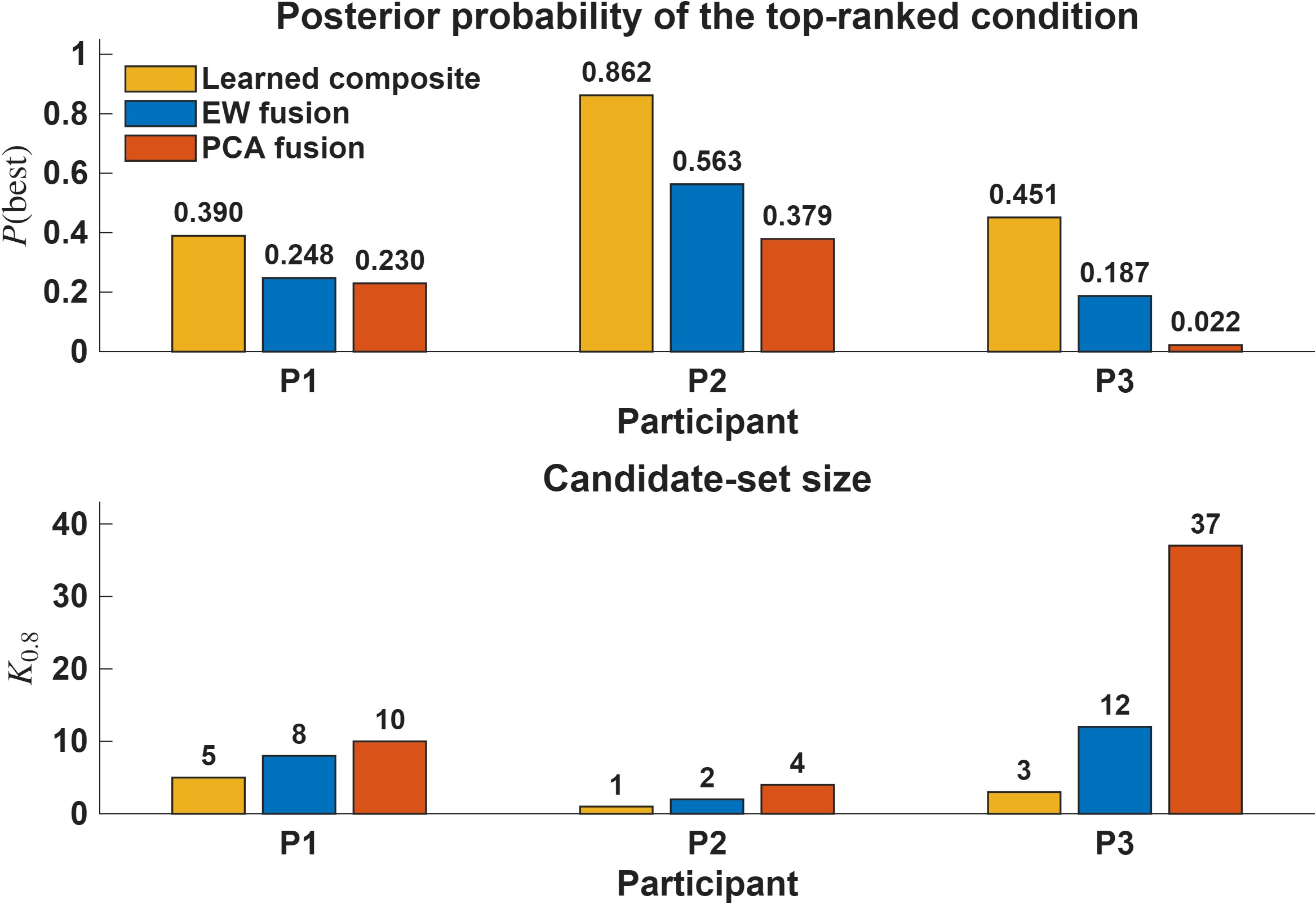}
  \vspace{-5mm}
\caption{Posterior decision summaries from the full-budget analysis ($B{=}4$ repeats per condition) for P1--P3. (A) Selected-condition $P(\mathrm{best})$. (B) Candidate-set size $K_{0.8}$, the minimum number of conditions reaching cumulative probability 0.8. Larger $P(\mathrm{best})$ and smaller $K_{0.8}$ indicate a more concentrated posterior ranking.}
\vspace{-5mm}
\label{figa}
\end{figure}

Fig.~\ref{figa} summarizes outputs from the full-budget analysis for the learned-composite selector and fusion baselines. For P1--P3, respectively, the learned-composite selector yielded $P(\mathrm{best})=0.390$, $0.862$, and $0.451$, with $K_{0.8}=5$, 1, and 3. EW fusion yielded $P(\mathrm{best})=0.248$, $0.563$, and $0.187$, with $K_{0.8}=8$, 2, and 12; PCA fusion yielded $0.230$, $0.379$, and $0.022$, with $K_{0.8}=10$, 4, and 37. The learned-composite selector had the largest $P(\mathrm{best})$ and smallest $K_{0.8}$ for every participant. EW fusion selected the same conditions, whereas PCA fusion differed for P3. Relative to EW fusion, the learned-composite selector concentrated probability within smaller candidate sets around the same selections, indicating lower posterior ambiguity. The no-torque reference comparison above provides exploratory biomechanical context.
\vspace{-3mm}

\subsection{Participant-specific weights and repeat-level stability}

\begin{table}[t]
  \centering
  \begin{threeparttable}
    \caption{Full-budget ($B=4$) learned fusion weights for the seven-sub-metric set}
    \label{tab:learned_weights}

    \scriptsize
    \setlength{\tabcolsep}{2.5pt}
    \renewcommand{\arraystretch}{1.05}

    \begin{tabular*}{\columnwidth}{@{\extracolsep{\fill}} lccc @{}}
      \toprule
      \textbf{Sub-Metric} & \textbf{P1} & \textbf{P2} & \textbf{P3} \\
      \midrule
      $\mathrm{MoS}_{\mathrm{post}}(t_2)$              & 0.233 & 0.431 & 0.051 \\
      peak~$|a_{y,\mathrm{CoM}}|$                      & 0.321 & 0.050 & 0.009 \\
      $\Delta x^{\max}_{\mathrm{CoM,post}}$           & 0.109 & 0.147 & 0.026 \\
      abs.~$\mathrm{iWBAM}_{\mathrm{sag}}$            & 0.145 & 0.002 & 0.003 \\
      peak~$|a_{\mathrm{AP},\mathrm{CoM}}|$           & 0.019 & 0.059 & 0.187 \\
      sag.~WBAM range change (\%)                     & 0.166 & 0.033 & 0.130 \\
      peak~$|\Delta y_{\mathrm{CoM}}|$                & 0.008 & 0.278 & 0.594 \\
      \midrule
      \textbf{Sum} & 1.000 & 1.000 & 1.000 \\
      \bottomrule
    \end{tabular*}
  \end{threeparttable}
  \vspace{-2mm}
\end{table}

Table~\ref{tab:learned_weights} reports the learned fusion weights. The largest weight was assigned to peak $|a_{y,\mathrm{CoM}}|$ for P1 (0.321), $\mathrm{MoS}_{\mathrm{post}}(t_2)$ for P2 (0.431), and peak $|\Delta y_{\mathrm{CoM}}|$ for P3 (0.594). For P1, five sub-metrics had weights between 0.109 and 0.321; for P2 and P3, the largest single weight accounted for 0.431 and 0.594 of the total, respectively. These patterns show that condition discrimination was supported by different sub-metric combinations across participants. Sampling variation can contribute to the observed weight differences, so their physiological interpretation remains participant-specific within the present dataset.

\begin{table}[t]
  \centering
  \begin{threeparttable}
    \caption{LORO refit stability and held-out ranking consistency}
    \label{tab:loro_stability}

    \scriptsize
    \setlength{\tabcolsep}{2.5pt}
    \renewcommand{\arraystretch}{1.05}

    \begin{tabular*}{\columnwidth}{@{\extracolsep{\fill}} lcccc @{}}
      \toprule
      \textbf{Participant} & $\overline{S}_w$ & \textbf{Selection} & \textbf{Retention} & $\overline{\rho}_{s}$ \\
      \midrule
      P1 & 0.874 & 2/4 & 3/4 & 0.324/0.361/0.318 \\
      P2 & 0.854 & 3/4 & 3/4 & 0.437/0.393/0.258 \\
      P3 & 0.911 & 2/4 & 4/4 & 0.249/0.062/-0.134 \\
      \bottomrule
    \end{tabular*}

    \begin{tablenotes}[flushleft]\scriptsize
      \item \textit{Notes:} $\overline{S}_w$ is the mean cosine similarity to the full-budget weights. Selection and Retention count folds that exactly select or retain the full-budget condition in $\mathcal{C}_{0.8}$. Spearman entries follow learned-composite/EW/PCA and are averaged across folds.
    \end{tablenotes}
  \end{threeparttable}
  \vspace{-5mm}
\end{table}

Table~\ref{tab:loro_stability} summarizes repeat-level stability. Mean weight cosine similarity ranged from 0.854 to 0.911. Full-budget selections were retained in 10 of 12 folds, and exact selection agreement occurred in 7 of 12 folds. Mean held-out Spearman correlations for the learned-composite selector were positive for all participants (0.249--0.437). These correlations exceeded PCA fusion for all participants and EW fusion for P2 and P3; EW was higher for P1. These results indicate moderate, participant-dependent stability. Lower selection agreement than retention shows that nearby conditions could exchange rank while remaining in the high-probability candidate set.

\section{LIMITATIONS AND FUTURE WORK}

The scope of these findings is defined by three healthy young adults, a discrete 46-condition assistance library, and four repeats per torque condition. The same trials included in the full-budget analysis supported weight learning, condition selection, and effect estimation, so the Welch comparisons are exploratory within-participant analyses with limited precision. Each LORO fold contained one held-out repeat per condition, which limited the precision of the held-out condition ranking and the fold-level stability summaries.

Randomized session and condition orders and interleaved no-torque reference trials distributed order effects across the protocol. Learning, fatigue, adaptation, and slow temporal variation may still have influenced the responses. The present dataset does not isolate these temporal contributions from torque-condition effects.

The homoscedastic EB hierarchical model pools within-condition variance. Condition-dependent variability could alter posterior uncertainty, $P(\mathrm{best})$, and candidate-set membership while leaving condition means unchanged. The sub-metric set emphasizes anterior--posterior perturbations and sagittal assistance. Mediolateral and step-level metrics, including step timing, step-width variability, and foot-placement predictability, remain outside the present scope and could redistribute the learned weights and condition rankings.

Future studies will combine larger, more diverse cohorts with independent sessions, explicit temporal-effect modeling, and moderated heteroscedastic EB estimation. Broader perturbation directions and step-level metrics will test how cost composition shapes selected assistance. These extensions will support safety-constrained online optimization over discrete and continuous torque parameterizations.

\section{CONCLUSION}
This study developed a participant-specific composite balance cost that integrates seven MoS-, CoM-, and WBAM-based sub-metrics, together with a learned-composite selector for ranking hip-exoskeleton assistance under limited repeats. Across three participants and 46 conditions, the method produced more concentrated posterior rankings than EW and PCA fusion. Selected-condition trials showed lower observed composite costs than no-torque trials, while LORO refits indicated positive held-out rank correlations and moderate stability. These proof-of-concept results support using the learned-composite selector to identify smaller high-probability candidate sets for follow-up testing in perturbation-based human-in-the-loop experiments.


\bibliographystyle{IEEEtran}
\bibliography{Master}

\end{document}